\documentclass[authoryear,preprint,12pt,3p]{elsarticle}
\pdfoutput=1

\usepackage{array}

\usepackage[greek,english]{babel}
\usepackage{float}
\usepackage[utf8]{inputenc}
\usepackage{amssymb}
\usepackage{xcolor,colortbl}
\usepackage{graphicx}
\usepackage[lofdepth,lotdepth]{subfig}
\usepackage{multirow}
\usepackage{setspace}
\usepackage{bm}
\usepackage{booktabs}
\usepackage{graphicx}
\usepackage[export]{adjustbox}
\usepackage{lipsum}
 \usepackage{lineno}
\usepackage{amsmath}
\usepackage{algorithm}
\usepackage[noend]{algpseudocode}

\usepackage{float}

\usepackage{hyperref}

\date{January 2019}

\journal{arXiv}

\begin{document}

\begin{frontmatter}

\title{
Improved prediction of soil properties with Multi-target Stacked Generalisation on EDXRF spectra}

\cortext[cor1]{Corresponding author}
\author[label1]{Everton Jose Santana}
\corref{cor1}
\ead{evertonsantana@uel.br}

\author[label2]{Felipe Rodrigues dos Santos}
\ead{fe.chicoo@gmail.com}

\author[label3]{Saulo Martiello Mastelini}
\ead{mastelini@usp.br}

\author[label2]{Fábio Luiz Melquiades}
\ead{fmelquiades@uel.br}

\author[label1]{Sylvio Barbon Jr.}
\ead{barbon@uel.br}

\address[label1]{Department of Computer Science,  State University of Londrina (UEL), Londrina, Brazil}

\address[label2]{Department of Physics, State University of Londrina (UEL), Londrina, Brazil}

\address[label3]{Institute of Mathematical Sciences and Computing, University of São Paulo (USP), São Carlos, Brazil}

\begin{abstract}
 Machine Learning (ML) algorithms have been used for assessing soil quality parameters along with non-destructive methodologies. Among spectroscopic analytical methodologies, energy dispersive X-ray fluorescence (EDXRF) is one of the more quick, environmentally friendly and less expensive when compared to conventional methods. However, some challenges in EDXRF spectral data analysis still demand more efficient methods capable of providing accurate outcomes. Using Multi-target Regression (MTR) methods, multiple parameters can be predicted, and also taking advantage of inter-correlated parameters the overall predictive performance can be improved. In this study, we proposed the Multi-target Stacked Generalisation (MTSG), a novel MTR method relying on learning from different regressors arranged in stacking structure for a boosted outcome. We compared MTSG and 5 MTR methods for predicting 10 parameters of soil fertility. Random Forest and Support Vector Machine (with linear and radial kernels) were used as learning algorithms embedded into each MTR method. Results showed the superiority of MTR methods over the Single-target Regression (the traditional ML method), reducing the predictive error for 5 parameters. Particularly, MTSG obtained the lowest error for phosphorus, total organic carbon and cation exchange capacity. When observing the relative performance of Support Vector Machine with a radial kernel, the prediction of base saturation percentage was improved in 19\%. Finally, the proposed method was able to reduce the average error from 0.67 (single-target) to 0.64 analysing all targets, representing a global improvement of 4.48\%.

\end{abstract}


\end{frontmatter}
\section{Introduction}

Evaluation of soil management is of fundamental importance in modern agriculture to achieve an effective soil correction with focus on highly productive crops as well as high harvesting performance \citep{suchithra2019improving}. 
In order to obtain such performance, precision agriculture with proximal soil sensor (PSS) is an optimal solution and a tendency. In this case, non-destructive spectroscopic analytical methodologies coupled to machine learning have been studied to correlate the analytical signal to the soil fertility parameter of interest \citep{wang2015soil, chlingaryan2018machine, mancini2019tracing, nawar2019can, rawal2019determination, zhang2019mapping, margenot2020predicting}. Particularly, X-ray Fluorescence (XRF) has been revealed as an upcoming PSS technique, especially with the popularisation of the portable XRF equipment \citep{sharma2015characterizing, dao2016instantaneous, morona2017quick, declercq2019comprehensive}.

The Energy Dispersive XRF (EDXRF) is the modality that has been successfully applied for soil parameters analysis in different fields such as agronomy and environment (soil pollution) \citep{weindorf2014advances}. The EDXRF spectral data signal or the soil elemental concentrations are obtained faster, it is non-destructive, environmentally friendly, and is less expensive than the conventional methods. These features make EDXRF feasible as a PSS. However, analytical drawbacks such as poor performance for low-Z elements, matrix effects (due to moisture, granulometry, complex soil composition) and spectral interferences are challenges to be overcome with improvement in data collection and data analysis. In this paper we focus on data analysis, i.e., in the use of machine learning algorithms to pursuit high performance regression models for soil fertility parameters. 

Commonly more than one parameter is involved in these analyses, forming an output set $Y$ composed of $d$ target variables. The traditional method to attack these problems is transforming the problem into $d$ sub-problems with a single output variable, sharing the same input set $X$. This method is known as Single-target (ST) and serves as a baseline when referring to problems with multiple targets. In ST, independent individual models (regressors) are generated for each sub-problem considering the same input set $X$ and the actual target.

Recent literature shows that new methods were developed especially to address multi-target settings. These methods, which are named Multi-target Regression (MTR) methods, oppose to ST by taking into account the correlation that the targets might have \citep{Borchani2015, Spyromitros2016, melki2017multi, Moyano2017, mastelini2018benchmarking}. Transforming the original problem into sub-problems is a well-grounded strategy of MTR based on two main strategies: stacking and chaining.

In stacking, one or more regressors are trained for each target (as in ST). They can be referred to as base-models. After that, predictions are obtained using the base-models for training a new regressor for each target in different manners. These new models, called meta-models, can be obtained following different stacking assumptions \citep{Spyromitros2016,Santana2017,sanatana2018}. The precursor was Stacked Single Target (SST) proposed by \cite{Spyromitros2016}. SST creates one base-model with a given regression algorithm (base-learner) for each target. Predictions are made by merging the output of base-models and the original input set, forming an augmented dataset. A new regressor is trained for each target taking into account the transformed dataset, generating $d$ meta-models in the second layer. \cite{Santana2017} proposed the Deep Regressor Stacking (DRS), which a stacking process of SST is repeated continuously for creating a deep meta-model. It stops when a maximum amount of pre-defined layers is reached. In another method, called Multi-target Augmented Stacking (MTAS), multiple distinct base-learners (related to different algorithms) are trained for each target using the same strategy as SST. However, the authors \citep{sanatana2018} took advantage of predictions for only relevant targets, obtaining boosted base-models. After that, just one final predictor per target is generated using the best set of augmented data, obtaining an accurate meta-model.

Chaining strategies, as the name implies, cascade the insertion of target-related information when creating new predictors. Similarly to stacking, chaining methods augment the original training set with predictions of the targets. Nonetheless, rather than using predictions of all responses at once, the chaining methods incrementally augment the datasets one target at the time. This idea is similar to the Bayesian network inference in design and was designed initially for classification problems~\citep{read2009classifier}. Different strategies were proposed in the recent MTR literature \citep{Spyromitros2016,Mastelini2018}.
The Ensemble of Regressor Chains (ERC) constructs multiple randomly ordered target chains~\citep{Spyromitros2016}. For each chain, base-models are trained for each target, starting from the first one. New regressors use the prediction of the previous base-models as extra input features. The final predictions for each target are accounted for as the average output between all chains. 
Multi-output Tree Chaining (MOTC)~\citep{Mastelini2018} constructs a tree structure rather than multiple chains, where each node represents a target. To this end, a measure of inter-target correlation is used. Once the tree is constructed, starting from the leaves to the root, MOTC starts training the base-models. Each node uses its descendants' base-models predictions as extra outputs. Thus, a specialised chain based on a tree branch is created to improve each target prediction.

The MTR methods have been applied to different fields, as to predict vegetation condition, water quality and rock mass parameters, wheat flour quality and even in soil assessment for heavy metal concentration \citep{Kocev2009,Spyromitros2016,junior2019multi,LIU2019}. The potential gains that MTR could bring motivated the application and evaluation of these methods to handle soil samples analysed by EDXRF, which presents difficulties of identification due the complexity of the soil EDXRF spectra.

Also, inspired by the stacking ensemble strategy, we developed a novel MTR method, Multi-target Stacked Generalisation (MTSG). Stacked Generalisation (SG) was the first stacking technique proposed in the literature, but it has not been explicitly addressed in the MTR tasks. In our proposal, as an ensemble technique, SG aims at combining the learning biases of different base-learners towards reducing the prediction error. MTSG follows the same principle, but differently from the other MTR stacking based methods, MTSG uses the original input set only in the first phase. After creating the first base-model and obtaining the predictions outputted by them, new base-models are created based only on those predictions as the input. Hence, MTSG not only models inter-target dependencies but also consider the learning strategies of different regressor algorithms to provide responses. As shown in the SG literature, if these regressors are dissimilar in their bias, this methodology is capable of reducing error~\citep{breiman1996stacked}. We believe this error reduction can also be achieved in the MTR field, particularly on EDXRF spectra, since the signal represents different properties and linearities from the same set of data. In other words, the proposed method is able to support the improved predictions of several targets taking advantage of different base-learner characteristics such as linearity, monotonicity and kernel function.

This work aims at evaluating MTR methods in an EDXRF soil spectra dataset to predict 10 soil quality parameters and attest the best predictor for each target. Besides, it assesses the performance of the new method MTSG for minimising the prediction error. Random Forest and Support Vector Machine (with linear and radial kernel) were used as base learners for the methods. 

After examining the background of this research, the work is organised as follows: Section \ref{sec_mtsg} introduces MTSG, the new developed multi-target method. Following this, Section \ref{sec_experiment} details the experimental setup, showing the dataset acquisition, the compared methods and algorithms, and the evaluation metrics. In Section \ref{sec_results}, the results and their discussion are presented. Section \ref{sec_conclusion} outlines the main conclusions of the work. Lastly, two appendices were added with descriptive statistic of the soil parameters and the MTSG performance in benchmarking datasets.


\section{Multi-Target Stacked Generalisation} 
\label{sec_mtsg}

Ensemble methods were used in different tasks to improve predictive performance over single predictors \citep{brown2005diversity, mendes2012ensemble}. The wisdom of ensembles consists of the combination of different specialised components (with local minima) to produce a global minimum. Besides, these components can be obtained by different learning processes (either by different learning algorithms, parameters or training sets). The mentioned components can be classifiers or regressors, depending on the kind of problem. Since we deal with a regression problem, the specialised components in this work equal to regressors.

Ensemble approaches evolve generally two steps: ensemble generation step and integration step. The first is when the components are built and in the second the output of the components are aggregated to generate a new prediction \citep{rooney2004random}. This aggregation can be made in different forms, for example, by a linear combination of the models.

\begin{figure}[H]
    \centering
    \includegraphics[width=0.8\textwidth]{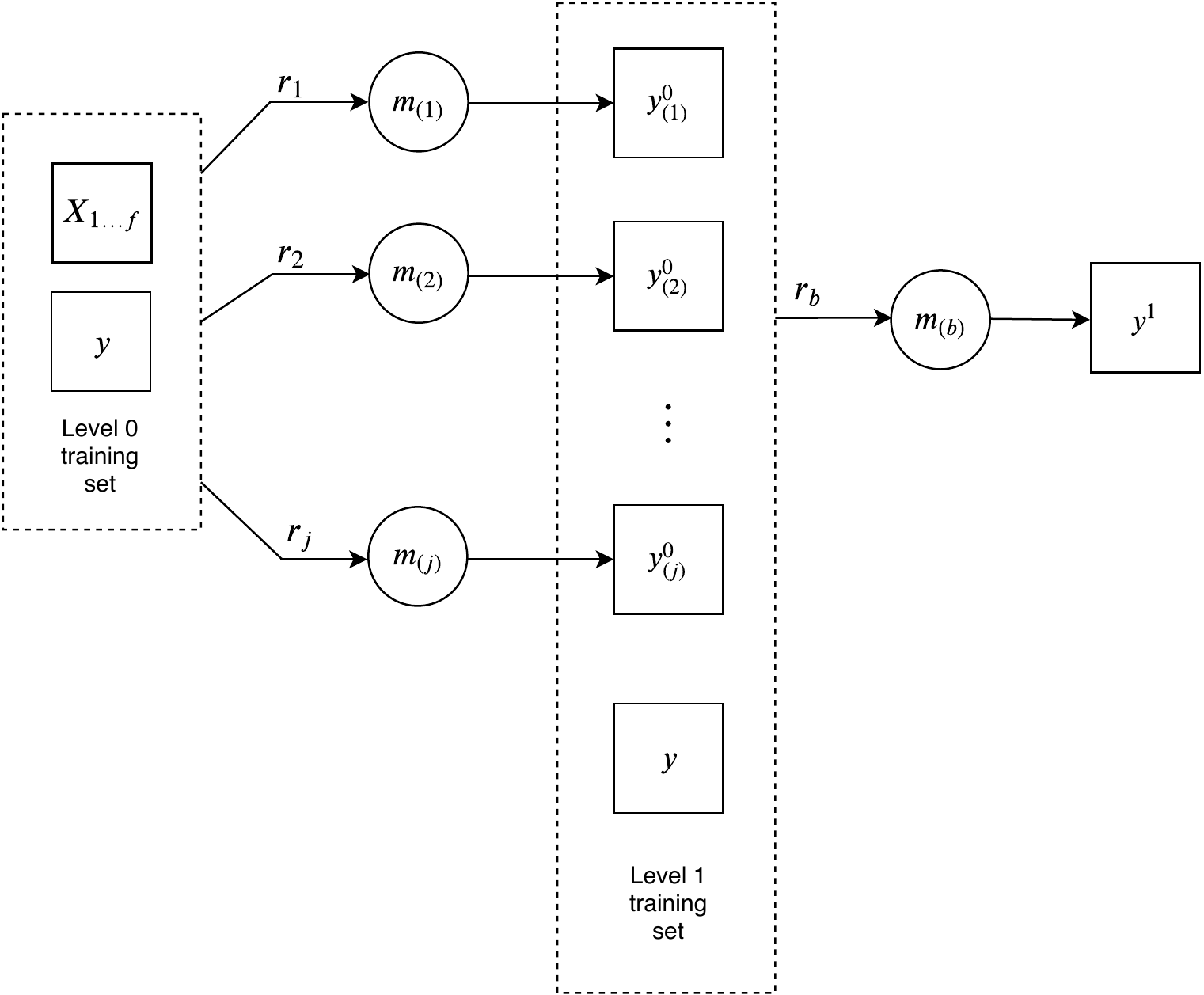}
    \caption{Representation of SG ensemble technique. }
    \label{fig_stacked}
\end{figure}

Another strategy to integrate the specialised components is to obtain a model that has as input the prediction generated by the components, process known as Stacked Generalisation (SG) \citep{wolpert1992stacked}. Figure \ref{fig_stacked} represents SG technique.

SG consists of two main phases. The Level 0 training set is composed of the input set $X$ with $f$ features and the output variable  $y$. Considering $R$ as the set of possible regression algorithms,  $j$ instances of it ($r_1 ... r_j \in R$) will be used to obtain the base-models $m_{(1)} \dots m_{(j)}$ for the targets. 

After creating these base-models, the $X$ set is used once again to obtain the first predictions ($Y^0$). The set of predicted targets $Y^0$ is then considered the new input set at Level 1.

In the second phase, one learning algorithm is chosen as the regressor ($r_b$). Thus, one Level 1 meta-model will be induced and considered the final predictor, and its prediction ($y^1$) will be considered the final output.

\begin{figure}[H]
    \centering
    \includegraphics[width=0.99\textwidth]{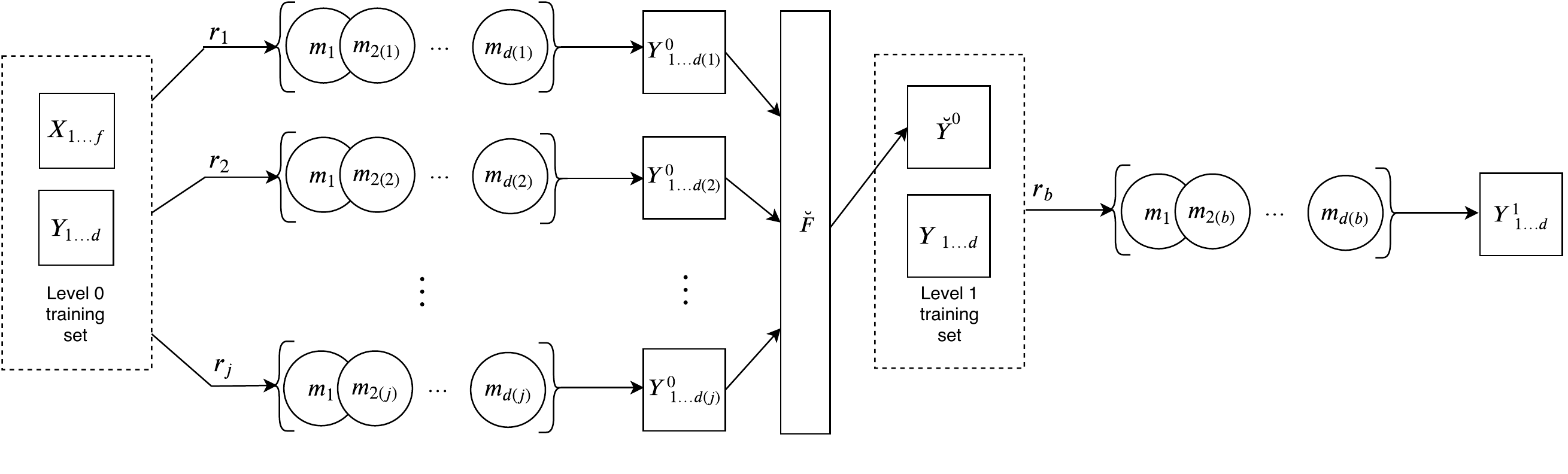}
    \caption{Representation of MTSG method.}
    \label{fig_mtsg}
\end{figure}

The Multi-Target Stacked Generalisation (MTSG) extends the SG concept to multiple outputs: whereas in original stacked generalisation multiple meta-models are trained for a single target, in MTSG multiple meta-models are trained for the multiple targets. Figure \ref{fig_mtsg} illustrates the MTSG design.

MTSG also has the generation and integration phases, with an intermediate pruning step made by a filter $\breve{F}$. The Level 0 training set is composed by the input set $X$ with $f$ features and the output set $Y$ with $d$ targets. $j$ base-learners will be used to obtain base-models for each target. In the first phase, $j$ base-models $m$ are induced for each target using the original training set.

The Level 0 base-models will be used along with the $X$ set to obtain the first predictions ($Y^0$). Since $j$ base-models are induced for each target at Level 0 and there are $d$ targets, $j \times d$ values will be delivered for each instance at this level. 

As an additional step, the set of predicted targets $Y^0$ will pass by the filter $\breve{F}$. This filter will assess the relevance of the predictions in relation to each target and will preserve only the relevant ones for each target ($\breve{Y^0}$). Those will be considered the new input set at Level 1.
For performing the filtering, different metrics can be used (e.g. liner correlation among the targets). In this work we adopted the importance extracted from Random Forest (RF), as in \cite{sanatana2018}, since it is capable of modelling nonlinear relationships.

In the integration step, similarly to SG, one learning algorithm is chosen as the base-learner ($r_b$). In a supervised fashion,  $d$ Level 1 meta-models will be induced, i.e., there will be one new regressor ($m_{(b)}$) for each target.  These  will be considered the final models and their predictions ($Y^1_{\ 1...d}$) will be considered the final output of the method.

The training procedure of MTSG is shown in Algorithm \ref{alg_MTSG}. It receives $X$ and $Y$, representing the input and output sets, respectively. $R$ represents the set of base-learners that will be used at Level 0 and $r_b$ represents the base-learner employed to create the final meta-models.

\begin{algorithm}[!h]
    \caption{MTSG training algorithm}
    \label{alg_MTSG}
    \begin{algorithmic}[1]
        \Function{MTSG}{$X$, $Y$, $R$, $r_b$}

            \State {\small // To store the Level 0 predictions}
            \State {$Y^0 \gets \{\}$}
            \State {\small // Level 0 base-models}
            \State {$\text{Level \ 0} \gets \{\}$}
            \State {\small // Regression model induction for each target and learner}
            \For{$r \in R$}
                \For {t = 1 \textbf{to} d}
                    \State {\small // model induction}
                    \State {$m_{t(r)}: X \xrightarrow{r} Y_{t}$}
                    \State {$Y^{0}_{t(r)} \gets \text{predict}(m_{t(r)}, X)$}
                    \State {$\text{Level 0}_{t(r)} \gets m_{t(r)}$}
                \EndFor
            \EndFor
            
            \State {\small // Second layer of meta-models}
            \State {$\text{Level}_{1} \gets \{\}$}
            \For {t = 1 \textbf{to} d}
                \State{$X^{'} \gets \breve{Y}^0_{\ t}$}
                \State {$m_{t(b)}: X^{'} \xrightarrow{r_b} Y_{t}$}
                \State {$\text{Level 1}_{{t}} \gets m_{t(b)}$}
            \EndFor
            \State {$mtsg \gets \{\text{Level 0, Level 1}\}$}
            
       \\ \Return $mtsg$
        \EndFunction
    \end{algorithmic}
\end{algorithm}

In this method, the original problem's features are disregarded at Level 1. To the best of our knowledge, this represents a distinction when comparing to the other multi-target stacking methods of the literature.

Besides, since it is an ensemble method, the performance of MTSG is sensitive to the diversity achieved by the Level 0 regressors. Using more learning algorithms tend to bring more diversity, however it increases the complexity of the method. For this reason, the number and type of regressors that will be used represent the compromise between performance and complexity in MTSG.

\section{Experimental Setup}\label{sec_experiment}
\subsection{EDXRF working principle}
EDXRF is a technique used in several study fields for identification and quantification of chemical elements present in varied materials \citep{byers2019xrf, khuder2010improvement, mantler2000x, parsons2013quantification, rodrigues2017evaluation, weindorf2012use}. The working principle of EDXRF depends on the interaction of high-energy X-rays with matter. These interactions may be performed by photoelectric effect, elastic and inelastic scattering \citep{van2002handbook}. The fluorescent phenomenon is related with the photoelectric effect in which electrons are ejected from inner shells of the atom by the incidence of an external X-ray beam. As a consequence, to stabilise the atom, electrons from external layers fill these vacancies and the energy difference is emitted as characteristic X-ray photons. Thus, it is possible to identify the elements present in the sample since these energies differences are well defined for each transition in each element. By evaluating the peak intensities in the spectra, the elemental concentrations may be calculated \citep{van2002handbook}. Besides the total inorganic content, the EDXRF spectra also provide some information about the samples complex organic content. The inorganic and organic samples information is mainly embodied in the spectra scattering region since the cross-section of the scattering effects is greater for low Z-elements.

\subsection{Conventional analyses and spectral measurements}

Soil samples (n=396) from an agricultural area in \textit{Ribeirão Vermelho} basin in Cambé municipality, Paraná State, Brazil (north region of Paraná state) were used in this study. The area has two types of soil classified as \textit{Latossolo Vermelho Amarelo Distrófico} (Orthic Ferralsol) and \textit{Nitossolo Vermelho Distrófico}, according to Brazilian classification system and FAO classification \citep{wrb2015world, santos2006sistema}, both with high clay texture. Samples were collected in three depths (0-5 cm, 5-10 cm e 10-20 cm), dried at 40 ºC for 48 h, grinded and sieved through a 2 mm stainless steel sieve. Next, soil samples were sent to laboratory for conventional analysis of the chemical parameters and EDXRF spectral measurements. 

The soil fertility parameters analysed were: bioavailable phosphorus (P), total organic carbon (TOC), pH, potential acidity ($H^+ + Al^{3+}$, that will be denoted by H+Al for simplification purpose), $Ca^{+2}$ (Ca), $Mg^{+2}$ (Mg), $K^+$ (K), sum of exchange bases (SB), cation exchange capacity (CEC) and base saturation percentage (BSP). All analyses were carried out in the IAPAR Soil Analysis Laboratory in Londrina, Paraná, Brazil following the recommendations of \cite{pavan1992manual}.

The EDXRF measurements were carried out in the Shimadzu (EDX720 model) benchtop equipment with Rh X-ray tube. For this, five grams of samples were placed in XRF plastic cups covered with Mylar films. The measurements were repeated three times in different sample portions (shaking the XRF cup before each measurement) using the  operation condition of 15 kV for 200 s. The detection was carried out using a Si (Li) detector cooled with liquid nitrogen. All samples were measured using 10-mm focal spot without any filter in the primary beam.

\subsection{Methods and algorithms}
The use of machine learning tools may help to work around and minimise the analytical drawbacks mentioned in the introduction and also perform rapid dataset analyses for soil characterisation \citep{kaniu2012direct, nawar2019can}.

The original dataset was split into two sub-sets following Kennard–Stone algorithm, which reserves 2/3 of the samples for training and 1/3 for test. This resulted in a training set containing 264 examples and a test set containing 132 examples.
The samples were further pre-processed using auto-scaling.

ST was compared to SST, ERC, MTAS, MOTC, DRS and the novel method, MTSG. Random Forest (RF) and Support Vector Machine (SVM) were used as Level 0 and base regressors in this work. All methods along with the base regressors were implemented in R 3.4.0 with default settings, and the implementation can be accessed in \footnote{\url{http://www.uel.br/grupo-pesquisa/remid/?page_id=145}}. The packages used for RF and SVM were \textit{ranger} and \textit{e1071}, respectively.
 
RF creates multiple decision trees considering subsets of training set features, forming a forest with specialised trees. The output of RF, when applied to regression problems, is the average of the trees in the forest.

SVM creates a hyperplane that minimises the training error. When using a linear kernel (SVM\_L), it creates a linear function to accomplish the task of minimising the error. When using a radial kernel (SVM\_R), it maps the training data to a higher dimension via a radial function and then finds a hyperplane that best minimises the error.

These algorithms were chosen due to the expressive performance in previous studies and to guarantee a diversity to ensemble, since they offer different biases.

\subsection{Evaluation metrics}

For analysing the quality of the methods concerning the prediction of each soil variable, the Root Mean Squared Error (RMSE) of each target $t$ was calculated between the predicted value ($y_t$) and the true value of the target ($\hat{y_t}$) for the $N$ testing instances. It indicates the concentration of the data in relation to the fitting model \citep{elavarasan2018forecasting}. 

\begin{equation}
   RMSE_t = \sqrt{\frac{ \sum_{i=1}^{N} (y_{t}^{i} -\hat{y}_{t}^{i})^2}{N}}
\end{equation}{}

Still focusing on the performance for each target, the multi-target methods can be compared to the single-target by the Relative Performance per Target (RPT), in which a value greater than 1 is a synonym of improved performance of MTR method in relation to the ST method:

\begin{equation}
    RPT_{MTR,t,r}=\frac{RMSE_{ST,t,r}}{RMSE_{MTR,t,r}}
\end{equation}

For this problem, $t$ corresponds to each soil parameter and $r$, to the base learner.

To compare the performance of different methods for different problems, it is necessary a metric that can be computed regardless of the number of targets $d$. The general performance of each method can be computed by the average Relative Root Mean Square Error (aRRMSE) \citep{Borchani2015}:
 
\begin{equation}
    aRRMSE_{MTR} = \frac{1}{d}\sum^{d}_{t=1} \sqrt{\frac{\sum_{i=1}^{N} (y_{t}^{i} - \hat{y}_{t}^{i})^2} {\sum_{i=1}^{N} (y_{t}^{i} - \overline{y})^2}}
    \label{eq_arrmse}
\end{equation}

A complementary metric to access the performance of the best models related to the reference values is the ratio of performance to deviation (RPD) which is the ratio of the standard deviation (SD) of the conventional analysis (reference) to the RMSE of the prediction.

\begin{equation}
    RPD = \frac{SD}{RMSE}
\end{equation}{}

According to \cite{rossel2006determining} for soil attributes RPD $<$ 1.0 indicates very poor predictions and their use is not recommended; RPD between 1.0 and 1.4 indicates poor predictions where only high and low values are distinguishable; RPD between 1.4 and 1.8 indicates fair predictions which may be used for assessment and correlation; RPD values between 1.8 and 2.0 indicates good predictions where quantitative predictions are possible; RPD between 2.0 and 2.5 indicates very good, quantitative predictions, and RPD $>$ 2.5 indicates excellent predictions. 

\section{Results and Discussion}
\label{sec_results}

The results were presented and discussed starting by exposing the correlation between the targets (Section~\ref{sub_corr}). This information corroborates with the importance of applying MTR methods since correlated targets lead to reduced errors in MTR predictions. Afterwards, in Section~\ref{sub_mtr}, we present the results of ST and MTR methods, highlighting the contribution of MTSG, which overcome the other methods. Finally, the predictions of each soil property are compared for evaluating their predictive performance and bring insights from patterns of EDXRF spectra in Section~\ref{sub_targets}.

\subsection{Soil properties correlation}
\label{sub_corr}

\begin{figure}[h]
    \centering
    \includegraphics[width=0.55\textwidth]{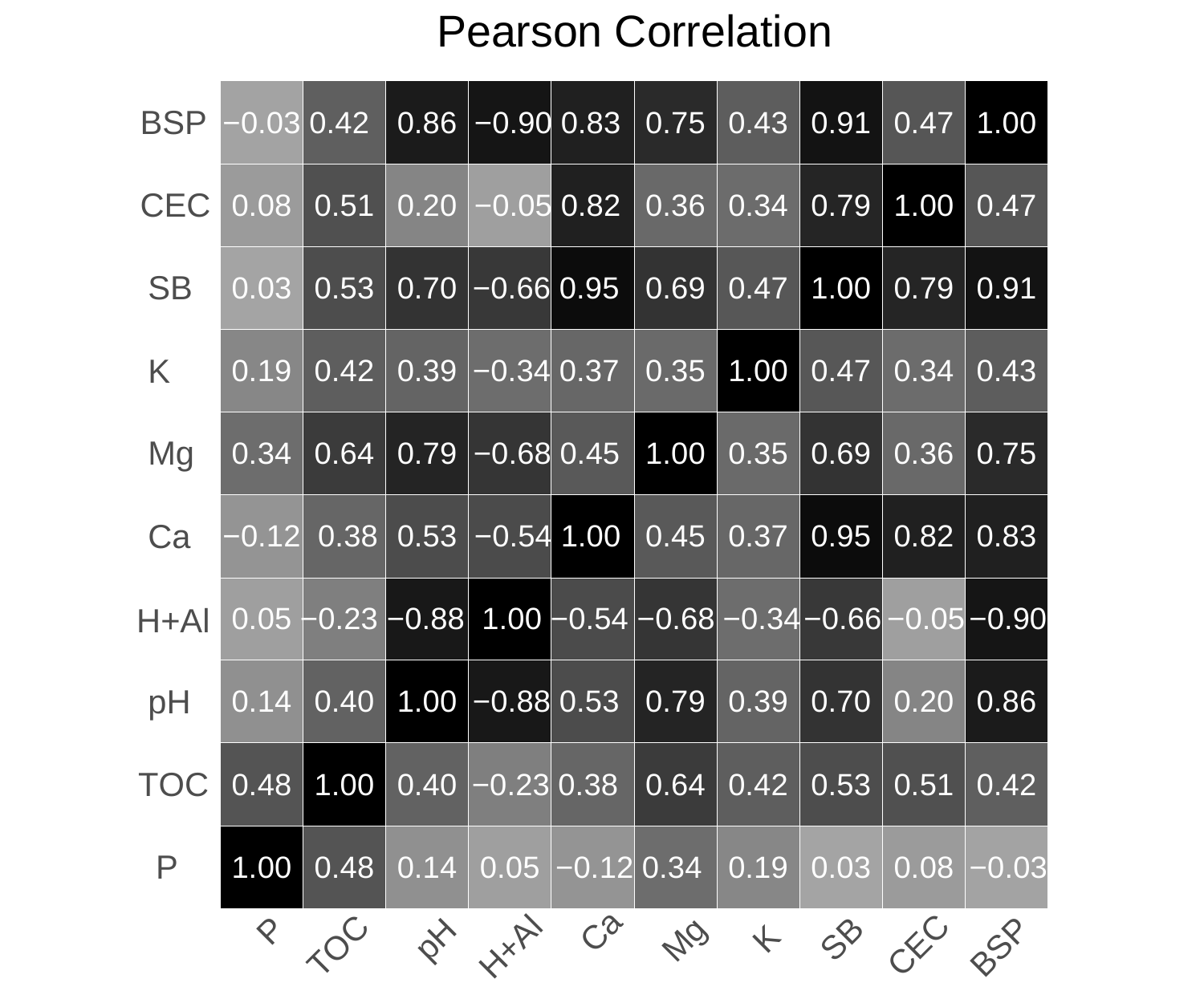}
    \caption{Pearson correlation coefficients among the 10 targets.}
    \label{fig_pearson}
\end{figure}

Figure \ref{fig_pearson} reveals the Pearson correlation coefficients among the targets. They were computed pair-wisely and the closer to 1 or -1, the higher the linear correlation among the two variables. Coefficients closer to 0 means a low correlation.

As observed, several targets are strongly correlated: Ca and SB result in the most expressive coefficient since they presented a correlation coefficient of 0.95, almost a perfect positive linear correlation. Other targets pairs that are strongly correlated are BSP and SB (0.91), BSP and pH (0.86), BSP and Ca (0.83), CEC and Ca (0.82), and Mg and pH (0.79). The targets BSP and H+Al presented a negative strong correlation (-0.90).  The targets that were mostly uncorrelated were P and SB (0.03), P and BSP (-0.03), P and H+Al (0.05), and P and CEC (0.08). In fact, P was the most uncorrelated target among the other nine.

\subsection{Multi-target prediction}
\label{sub_mtr}

\begin{figure}[ht!]
    \centering
    \includegraphics[width=\textwidth]{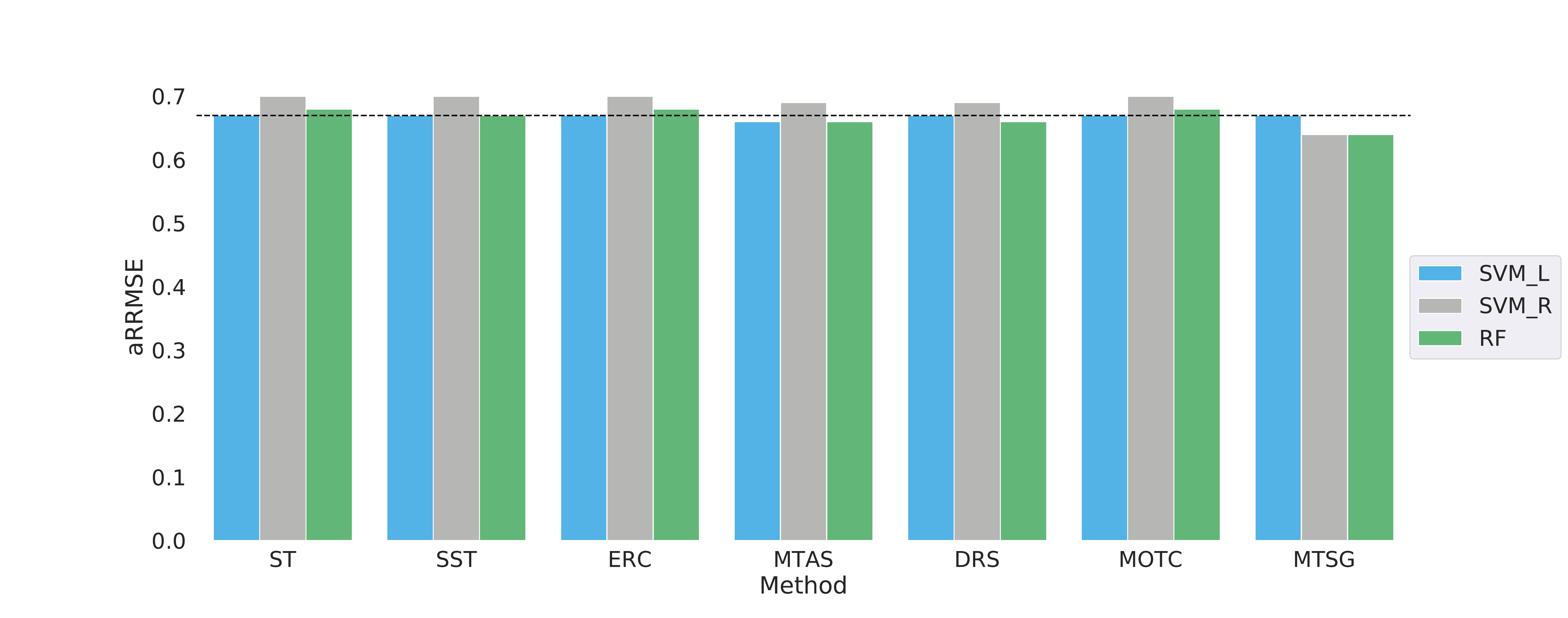}
    \caption{aRRMSE for the different combinations of methods and regressors. The horizontal line represents the lowest aRRMSE for ST.}
    \label{fig_res-geral}
\end{figure}

Figure \ref{fig_res-geral} shows the aRRMSE for all methods and base-learners. For this, the lower the value, the better the performance. Comparing the base-learners with the average error of all targets, SVM\_L was an effective machine learning algorithm, obtaining the same performance in all methods (ST and MTR, except for MTAS). SVR\_R and RF were able to reduce the average error only when used embedded in MTAS and MTSG.

{}

With ST, the lowest aRRMSE was obtained using SVM with a linear kernel, resulting in an aRRMSE of 0.67. Despite of introducing non linearities, the MTSG presented a significative improvement in the predictive power of the models.  MTSG using SVM with a radial kernel and RF as base regressor led to the lowest aRRMSE, 0.64. Other improvements in relation to the lowest ST value were obtained by DRS using RF (0.66), MTAS with SVM\_L (0.66) and MTAS with RF (0.66).


After evaluating the general error results of the methods, their performances were evaluated for each target in terms of RPT, as registered in Table \ref{tab_RPT}. For this metric, values greater than 1 represent an improvement of MTR over ST.

For P, TOC, pH, H+Al, Ca, Mg, K, SB, CEC and BSP, i.e., all the targets, all the combinations of MTR methods obtained improved or at least equivalent performance with respect to their corresponding ST version. It is noteworthy that many improvements reached up to around 10\% (using MTSG) in relation to the corresponding ST. 
When observing the average RPT (last column in Table \ref{tab_RPT}), just MTAS was able to improve results using SVM\_L, the other MTR methods shared the same performance of ST for this algorithm. However, using SVM\_R and RF, MTSG was the most promising method.
It may also be noted in Table 1 that RF generated a performance improvement in all the MTRs in relation to ST for SB.


\begin{table}[h!]
    \centering
\resizebox{0.8\textwidth}{!}{
\begin{tabular}{c|c|cccccccccc|c}
\cline{2-13}
& Regressor & P & TOC & pH & H+Al & Ca & Mg & K &SB& CEC & BSP & Average\\
\hline
$RPT_{SST}$ & SVM\_L & 1.00 & 1.00 & 1.00 & 1.00 & 1.00 & 1.00 & 1.00 & 1.00 & 1.00 & 1.00 & 1.00    \\
$RPT_{SST}$ & SVM\_R & 1.00 & 1.04 & 1.00 & 1.00 & 1.00 & 1.00 & 1.00 & 1.00 & 1.00 & 1.01 & 1.01    \\
$RPT_{SST}$ & RF     & 1.01 & 1.00 & 1.00 & 1.00 & 1.13 & 1.00 & 1.00 & 1.10 & 1.00 & 1.03 & 1.03    \\\hline   
$RPT_{ERC}$ & SVM\_L & 1.00 & 1.00 & 1.00 & 1.00 & 1.00 & 1.00 & 1.00 & 1.00 & 1.00 & 1.00 & 1.00    \\
$RPT_{ERC}$ & SVM\_R & 1.00 & 1.04 & 1.00 & 1.00 & 1.00 & 1.00 & 1.00 & 1.00 & 1.00 & 1.00 & 1.00    \\
$RPT_{ERC}$ & RF     & 1.01 & 1.00 & 1.00 & 1.00 & 1.13 & 1.00 & 1.00 & 1.10 & 1.00 & 1.02 & 1.03    \\\hline   
$RPT_{MTAS}$ & SVM\_L & 1.01 & 1.05 & 1.00 & 1.00 & 1.00 & 1.00 & 1.00 & 1.00 & 1.13 & 1.03 & \textbf{1.02}    \\
$RPT_{MTAS}$ & SVM\_R & 1.01 & 1.04 & 1.00 & 1.00 & 1.00 & 1.00 & 1.00 & 1.00 & 1.00 & 1.04 & 1.01    \\
$RPT_{MTAS}$ & RF     & 1.03 & 1.04 & 1.00 & 1.00 & 1.13 & 1.00 & 1.00 & 1.10 & 1.00 & 1.03 & 1.03    \\\hline   
$RPT_{MOTC}$ & SVM\_L & 1.00 & 1.00 & 1.00 & 1.00 & 1.00 & 1.00 & 1.00 & 1.00 & 1.00 & 1.00 & 1.00    \\
$RPT_{MOTC}$ & SVM\_R & 1.00 & 1.04 & 1.00 & 1.00 & 1.00 & 1.00 & 1.00 & 1.00 & 1.00 & 1.00 & 1.00    \\
$RPT_{MOTC}$ & RF     & 1.00 & 1.00 & 1.00 & 1.00 & 1.00 & 1.00 & 1.00 & 1.10 & 1.00 & 1.03 & 1.01    \\\hline   
$RPT_{DRS}$ & SVM\_L & 1.00 & 1.00 & 1.00 & 1.00 & 1.00 & 1.00 & 1.00 & 1.00 & 1.00 & 1.00 & 1.00    \\
$RPT_{DRS}$ & SVM\_R & 1.01 & 1.04 & 1.00 & 1.00 & 1.00 & 1.00 & 1.00 & 1.00 & 1.00 & 1.04 & 1.01    \\
$RPT_{DRS}$ & RF     & 1.01 & 1.00 & 1.00 & 1.00 & 1.13 & 1.00 & 1.00 & 1.10 & 1.00 & 1.08 & 1.03    \\ \hline   
$RPT_{MTSG}$ & SVM\_L & 1.00 & 1.00 & 1.00 & 1.00 & 1.00 & 1.00 & 1.00 & 1.00 & 1.00 & 1.00 & 1.00    \\
$RPT_{MTSG}$ & SVM\_R & 1.06 & 1.14 & 1.00 & 1.11 & 1.13 & 1.00 & 1.11 & 1.10 & 1.13 & 1.19 & \textbf{1.10}    \\
$RPT_{MTSG}$ & RF     & 1.08 & 1.14 & 1.00 & 1.00 & 1.13 & 1.00 & 1.00 & 1.10 & 1.13 & 1.00 & \textbf{1.06}    \\
\hline
\end{tabular}}
    \caption{RPT for each target considering the different combinations of methods and regressors. The best average result of regressor are highlighted by bold.}
    \label{tab_RPT}
\end{table}

\subsection{Soil properties prediction}
\label{sub_targets}

\begin{figure*}[]
\footnotesize
   \centering
\begin{tabular}{cc}
\footnotesize
\includegraphics[width=0.46\textwidth]{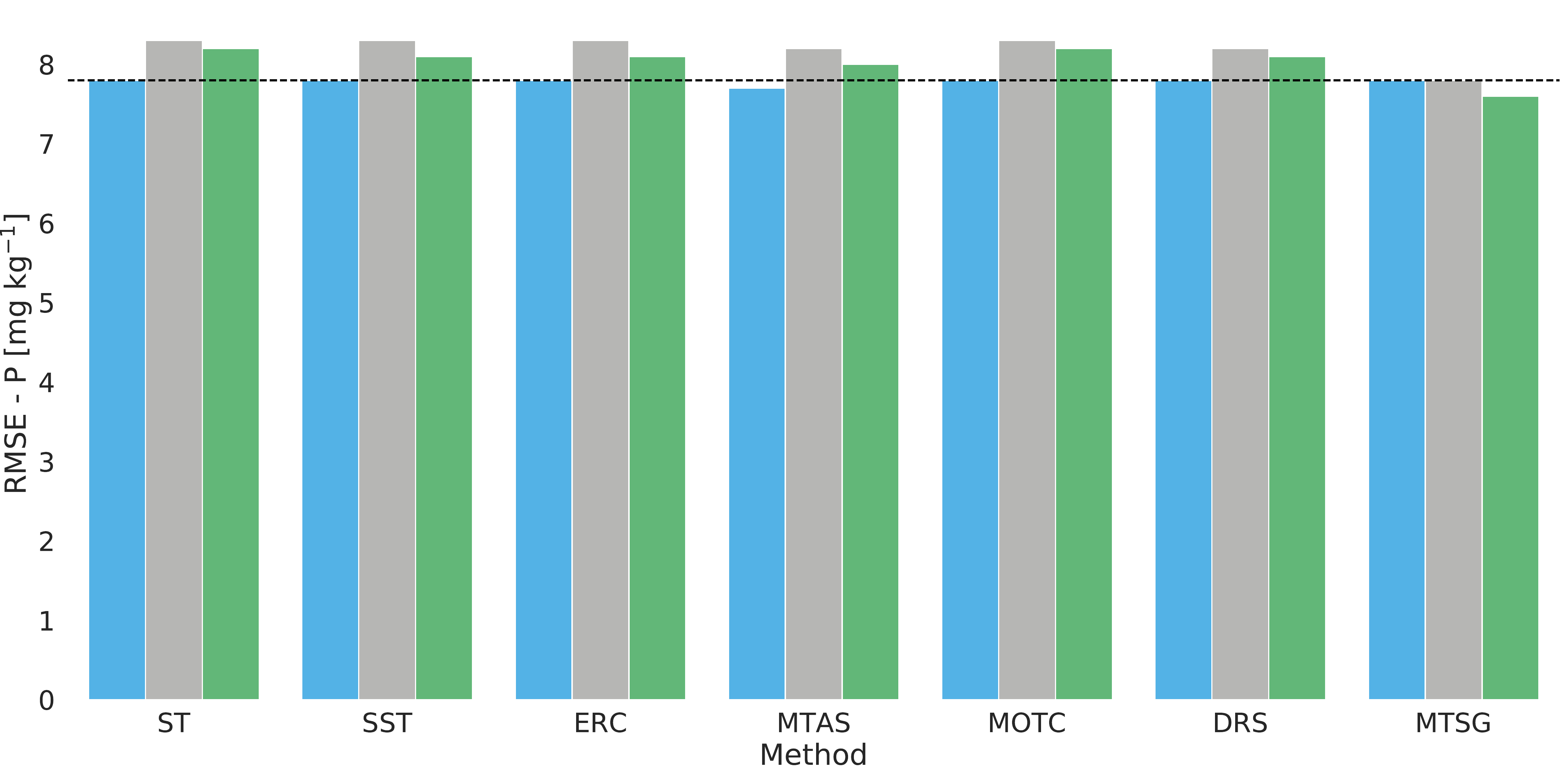}&
\includegraphics[width=0.46\textwidth]{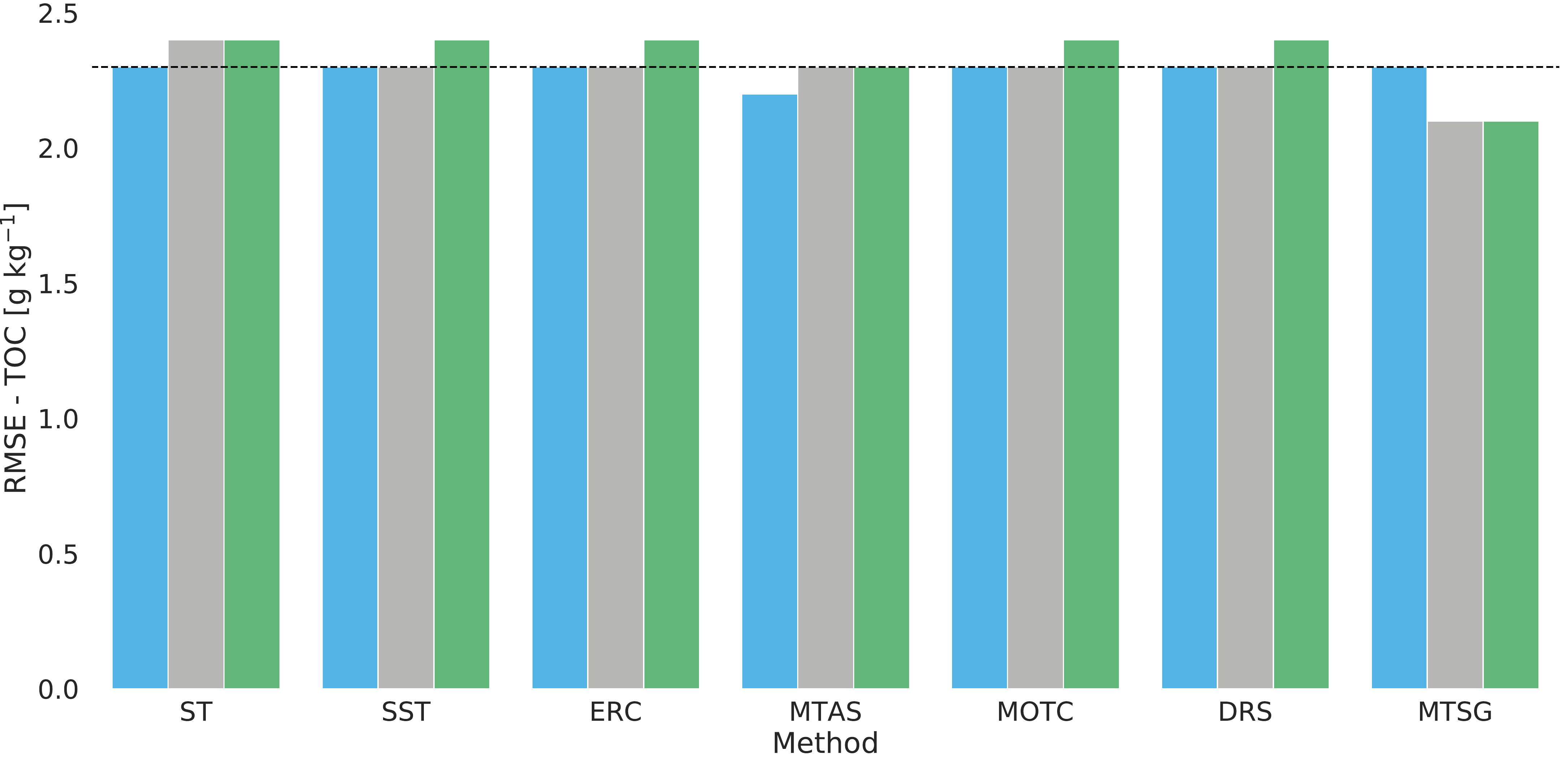}\\
a) P & b) TOC\\
\includegraphics[width=0.46\textwidth]{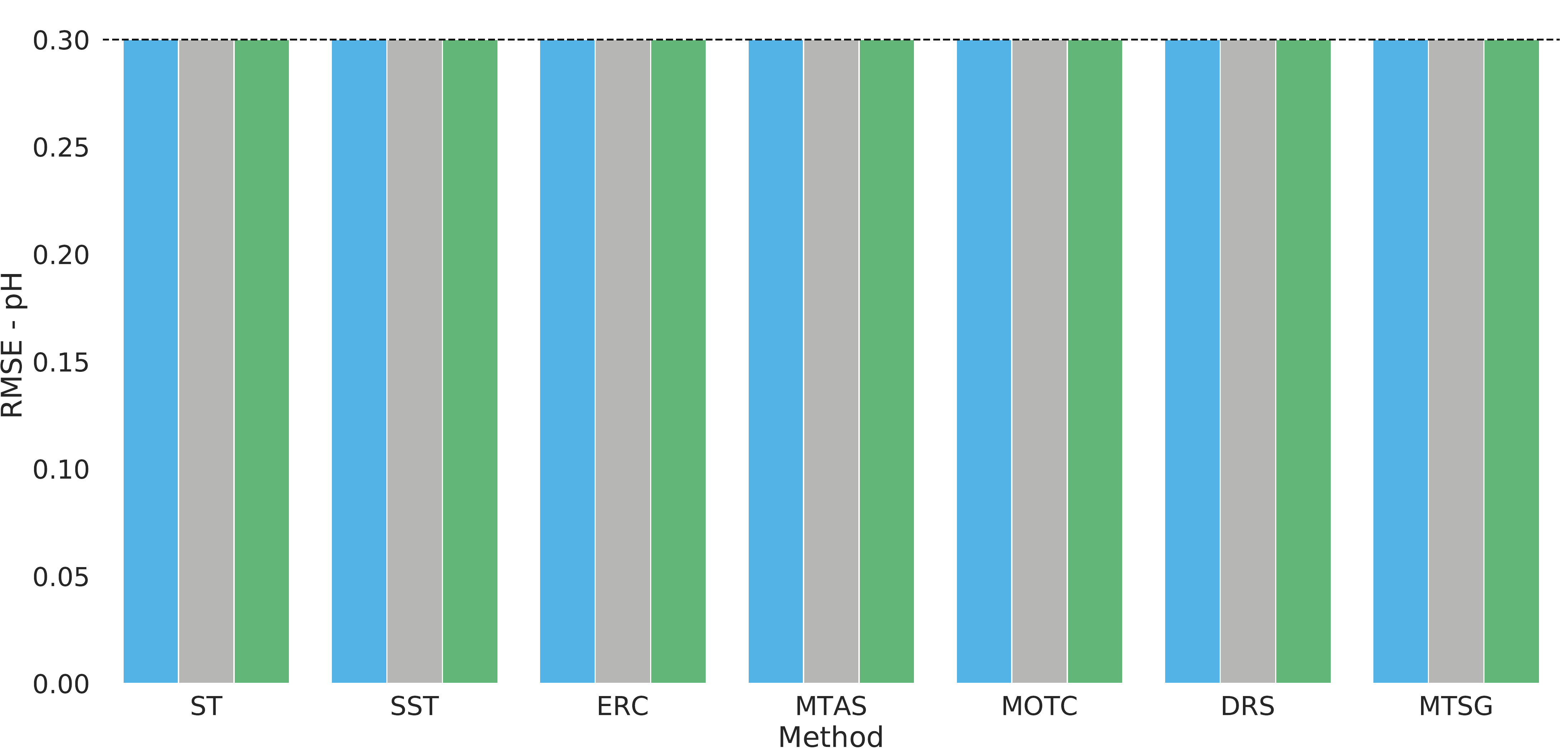}&
\includegraphics[width=0.46\textwidth]{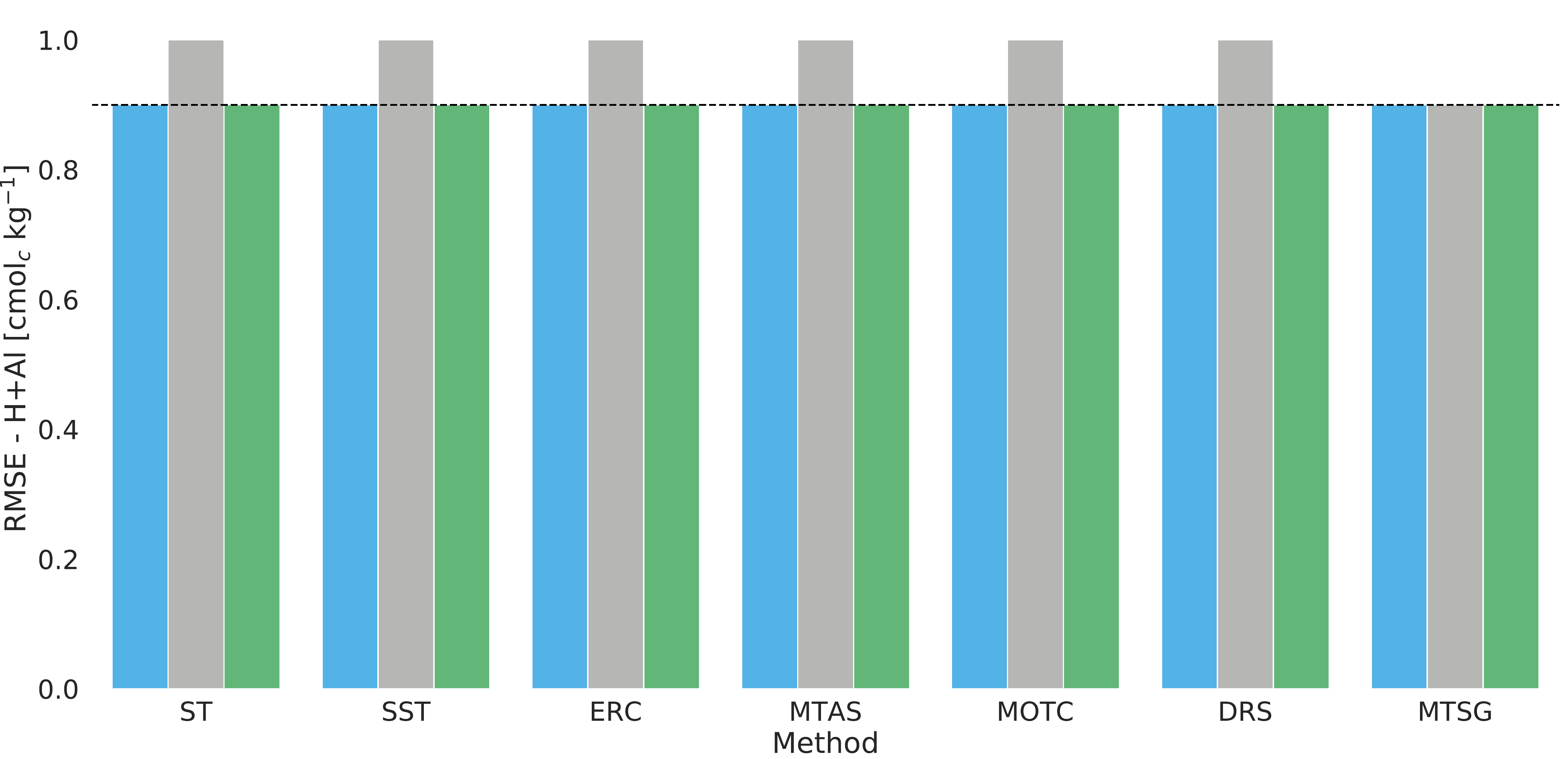}\\
c) pH & d) H+Al\\
\includegraphics[width=0.46\textwidth]{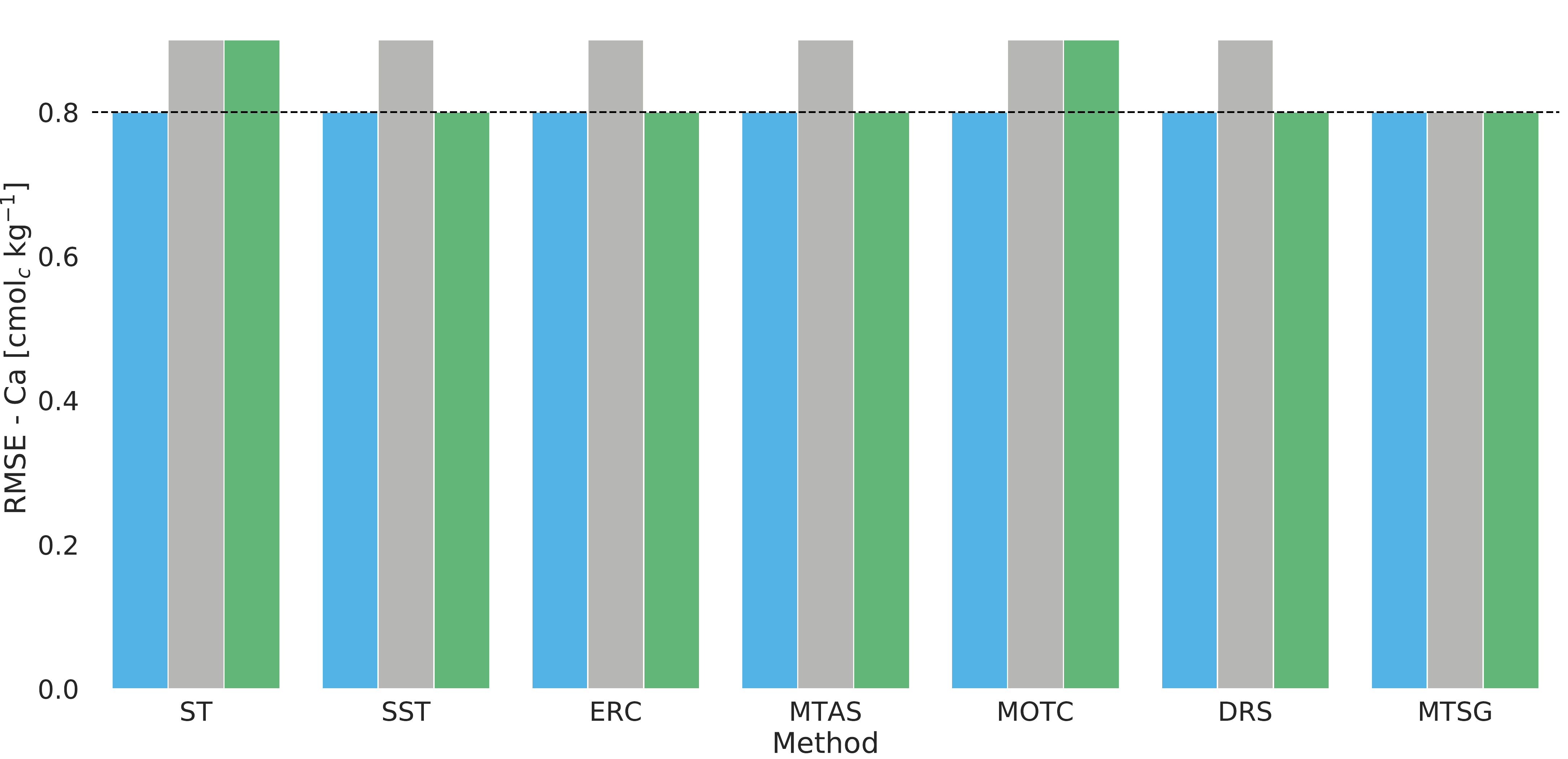}&
\includegraphics[width=0.46\textwidth]{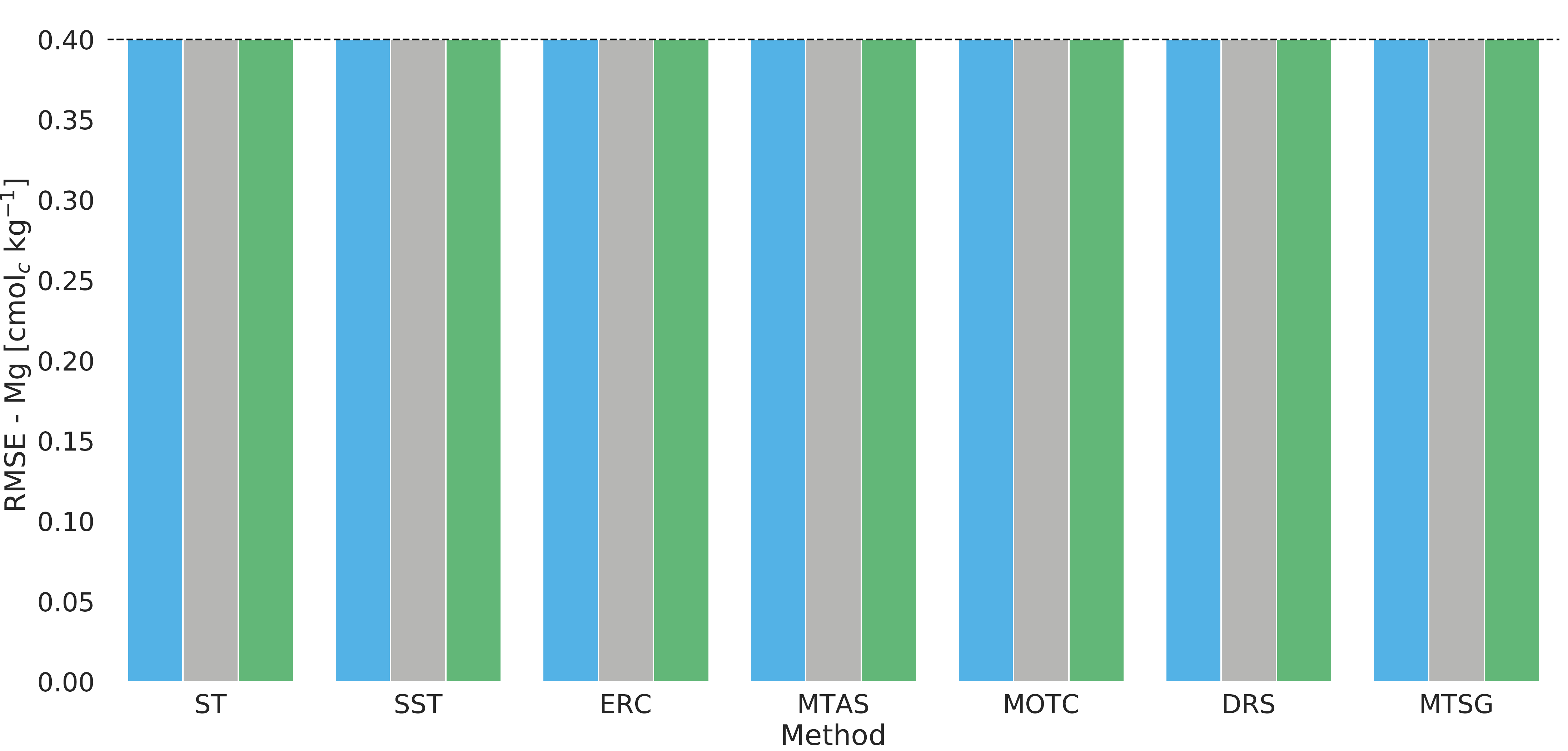}\\
e) Ca & f) Mg\\
\includegraphics[width=0.46\textwidth]{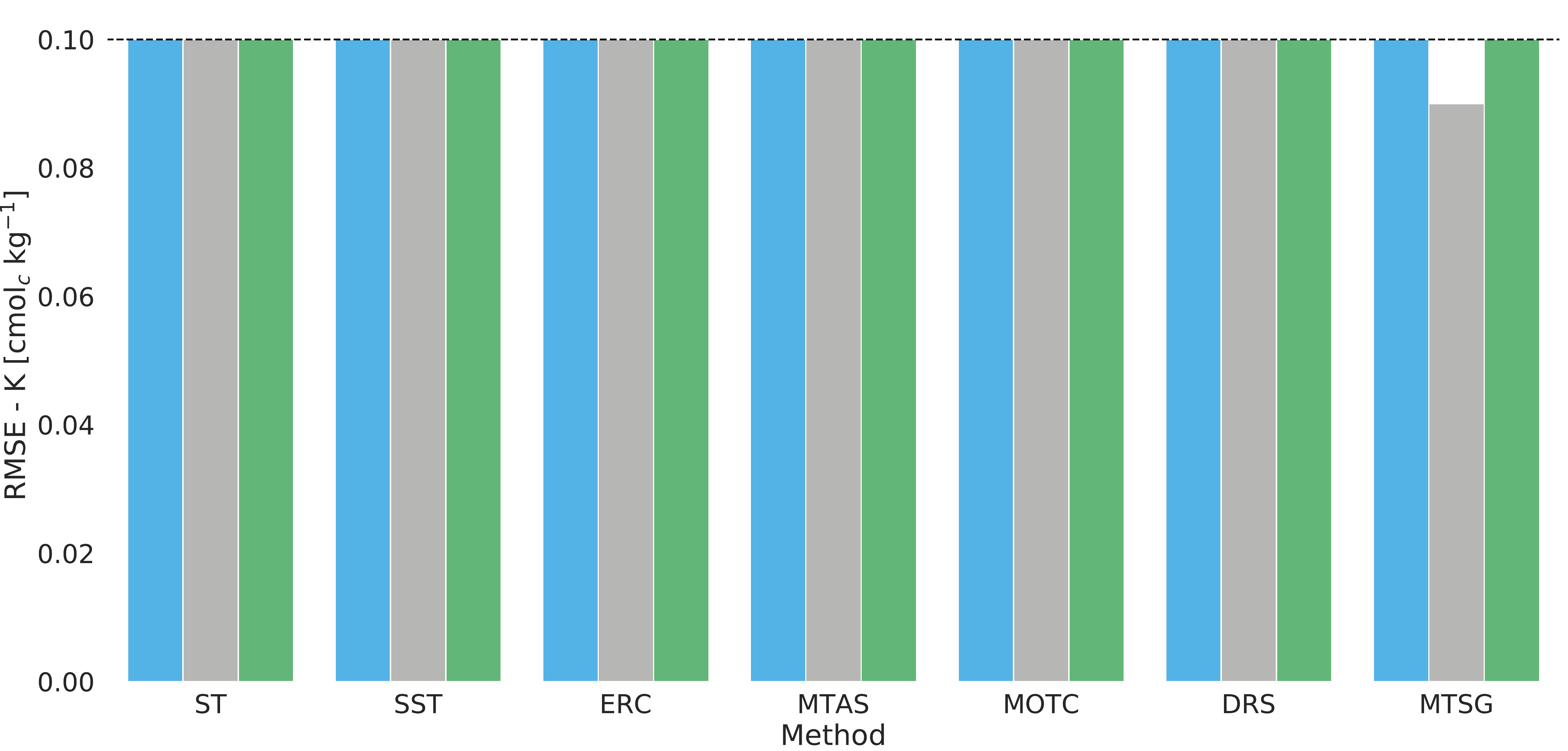}&
\includegraphics[width=0.46\textwidth]{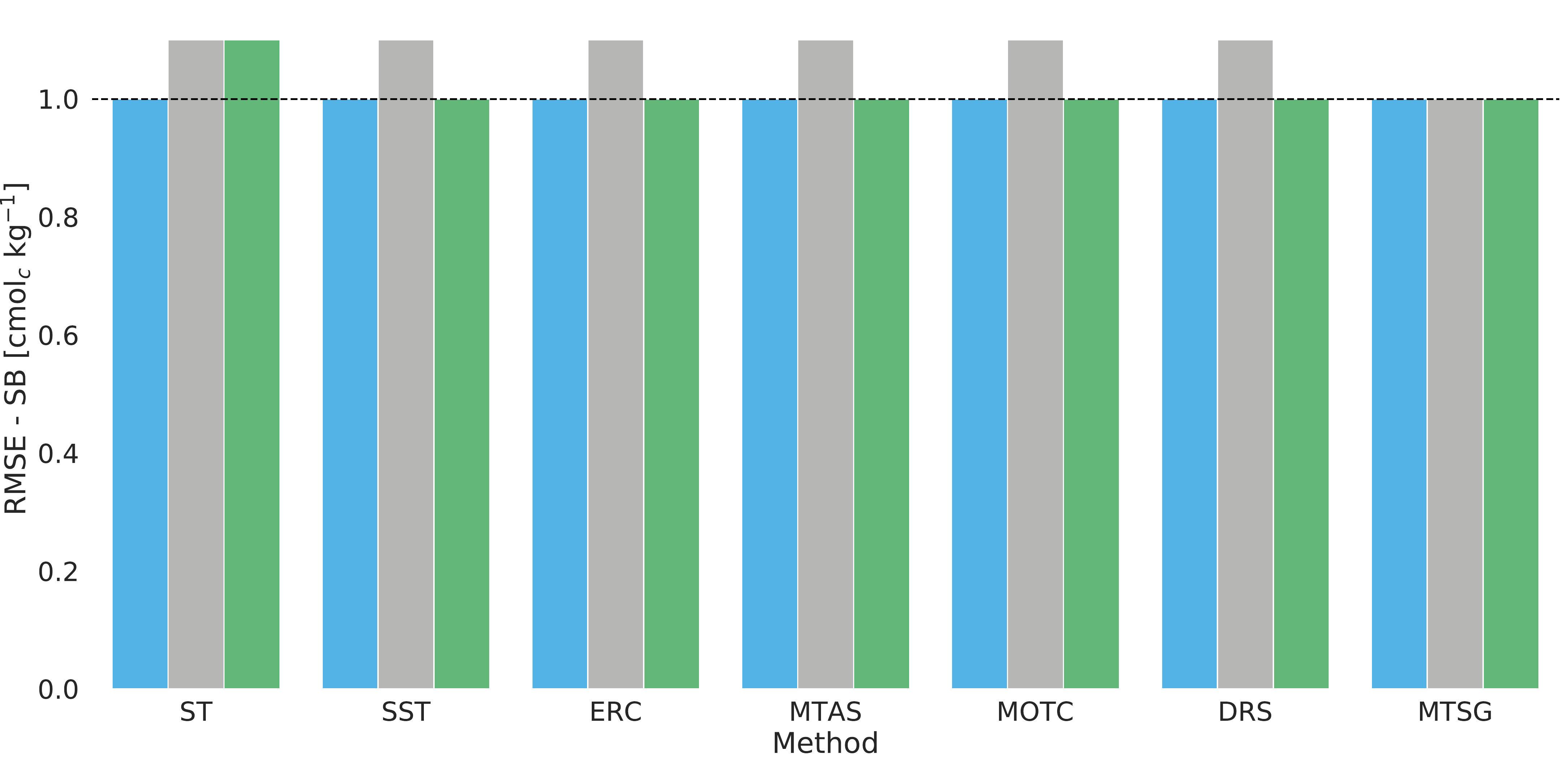}\\
g) K & h) SB\\
\includegraphics[width=0.46\textwidth]{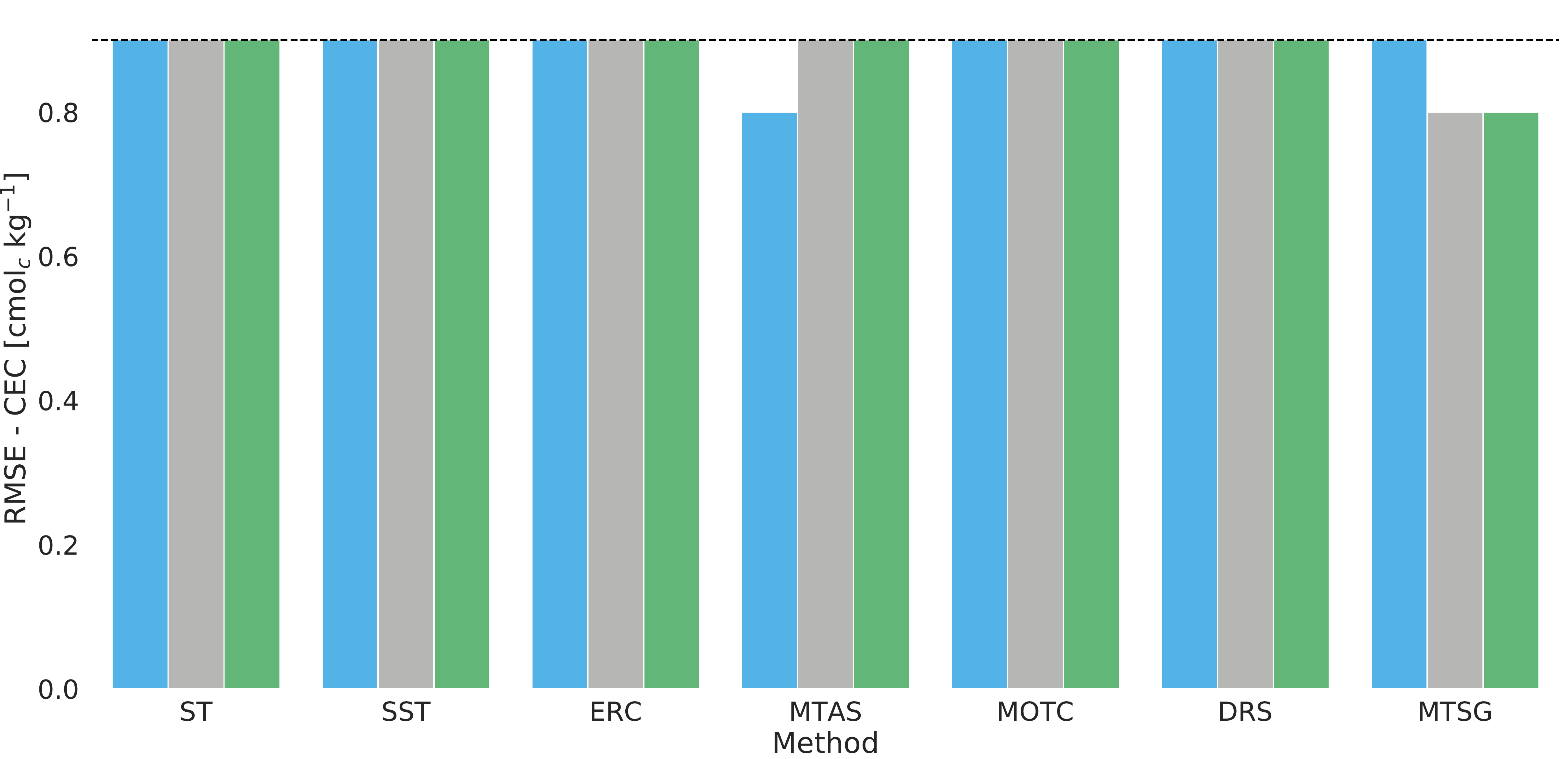}&
\includegraphics[width=0.46\textwidth]{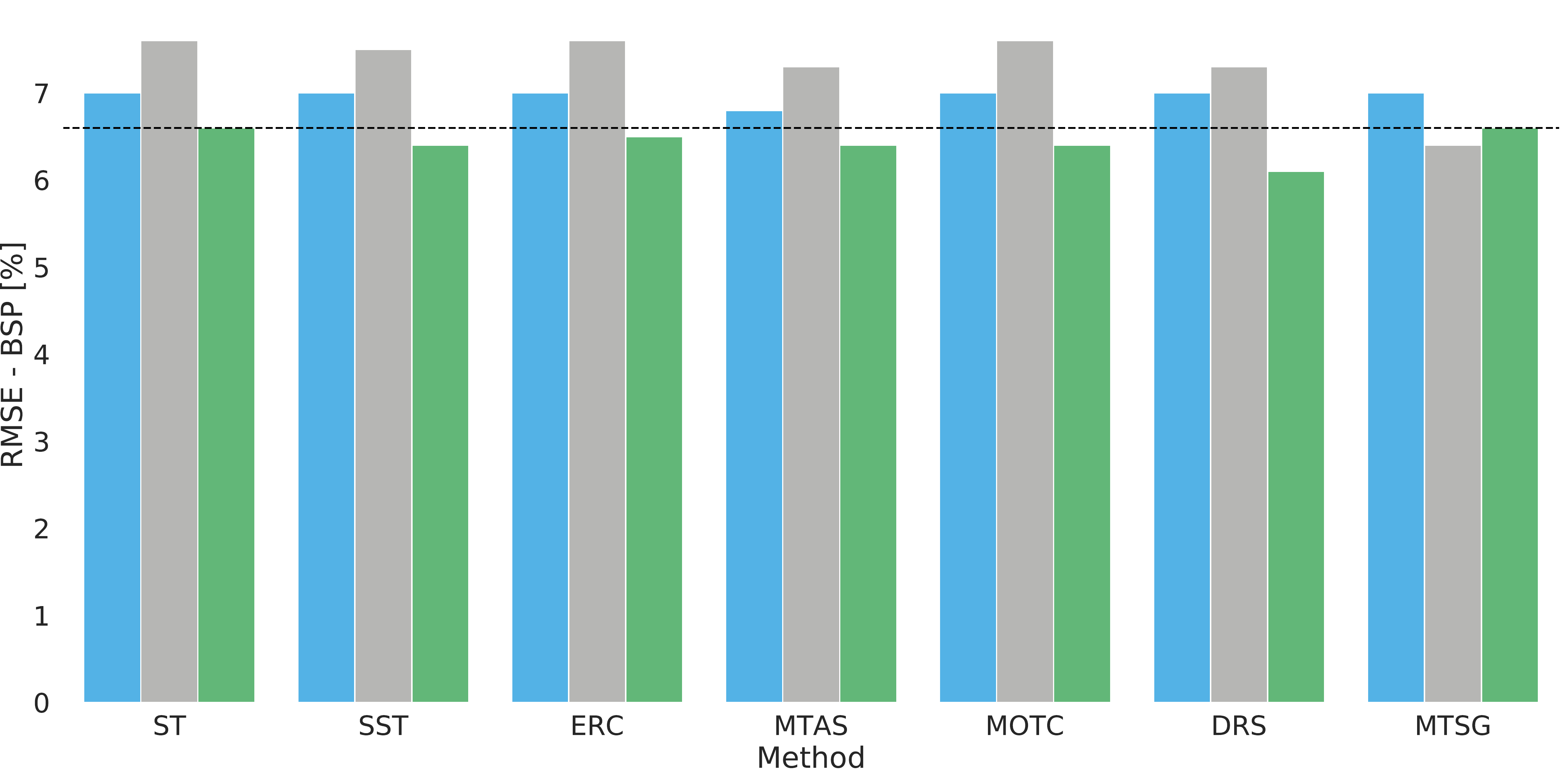}\\
i) CEC & j) BSP\\
\multicolumn{2}{c}{ \includegraphics[width=0.46\textwidth]{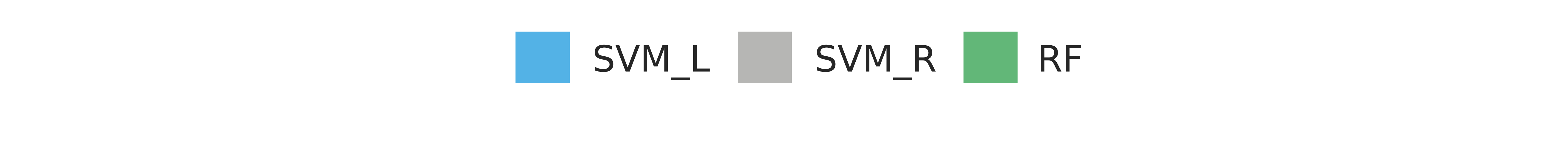}}\\
\end{tabular}
    \caption{RMSE values for each target considering the different combinations of methods and base-learners. The horizontal lines represent the lowest RMSE for ST related to each target.}
    \label{fig_res-RMSE} 
\end{figure*}

Continuing the discussion for individual targets, we present (Figure \ref{fig_res-RMSE}) the absolute error values between the ST and the MTR methods for each target.   For all the targets, except for BSP, ST models presented the lowest RMSE using SVM\_L, and thus indicates a linear relationship between the EDXRF spectral data and the studied targets.

Two properties, pH and Mg, did not improve their predictive performances through the use of MTR. For pH and Mg, all base-learners (SVM\_L, SVM\_R and RF) provided the same lowest RMSE in ST and MTR methods, respectively 0.3 and 0.4 cmol$_c$ kg$^{-1}$. 

All the other properties, we have at least one meta-model with reduced error in comparison to ST. Regarding K, all ST and MTR methods obtained 0.10 cmol$_c$ kg$^{-1}$, except our proposal. MTSG (SVM\_R) was able to reduce the error to 0.09 cmol$_c$ kg$^{-1}$. For P, SVM\_L provided the lowest RMSE in ST method (7.8 mg kg$^{-1}$). Considering P and MTR methods, MTSG with RF was able to reduce this error to 7.6 mg kg$^{-1}$. For TOC,  SVM\_L provided the lowest RMSE in ST method (2.3 g kg$^{-1}$). Regarding TOC and MTR methods, MTSG with SVM\_R and RF was able to reduce the error to 2.1 g kg$^{-1}$. Dealing with H+Al, SVM\_L and RF provided the lowest RMSE in ST method (0.9 cmol$_c$ kg$^{-1}$). About MTR methods and this property, SVM\_L and RF achieved the same error (0.9 cmol$_c$ kg$^{-1}$). MTSG with SVM\_R was also able to achieve this value. When predicting Ca, SVM\_L provided the lowest RMSE in ST method (0.8 cmol$_c$ kg$^{-1}$). Using MTR methods, SST, ERC, MTAS and DRS with SVM\_L and RF achieved the same error (0.8 cmol$_c$ kg$^{-1}$). MOTC achieved this error only using SVM\_L and MTSG achieved this error with RF, SVM\_L and SVM\_R. Regarding SB, SVM\_L provided the lowest RMSE in ST method (1.0 cmol$_c$ kg$^{-1}$). Considering MTR methods, SVM\_L and RF achieved the same error (1.0 cmol$_c$ kg$^{-1}$). MTSG with SVM\_R was also able to achieve this value. For CEC, SVM\_L, SVM\_R and RF provided the lowest RMSE in ST method (0.9 cmol$_c$ kg$^{-1}$). Considering MTR methods, MTAS with SVM\_L and MTSG with SVM\_R and RF were able to reduce the error to 0.8 cmol$_c$ kg$^{-1}$. Finally, regarding BSP, RF provided the lowest RMSE in ST method (6.6\%). Considering MTR methods, DRS with RF was able to reduce the error to 6.1\%.

To summarise, in ST method, for pH and, Mg, all base-learners shared the same RMSE. RF alone presented the lowest error in ST only for BSP. Thus, for ST method, SVM\_R was not the unique best regressor for any of the targets, and there is a prevalence of SVM\_L. This scenario, however, changed when these regressors were used as base-learners for the MTR methods: ignoring the cases that MTR did not bring improvements to ST, RF was present as the base-learner for 4 targets (P, TOC, CEC and BSP), SVM\_R for 3 targets (TOC, K and CEC) and SVM\_L for 1 target (CEC). This shows that the modifications inherent to MTR methods introduced non-linearity to the data. 

P is one target that was favoured by this non-linearly: the Pearson correlation coefficients of P with other targets, as discussed in Figure \ref{fig_pearson}, revealed that it was least linearly correlated to other targets. Nonetheless, its RMSE lowered with MTSG coupled with RF.

As observed, the best MTR methods resulted in RMSE values at least equivalent to ST. For the targets pH, H+Al, Ca,  Mg and SB, none of the MTR methods improved the performance, so ST would be preferable to predict them due to the lower complexity. On the other hand, for the other 5 targets, at least one MTR method improved the performance, specially MTSG, but also DRS and MTAS.

In order to evaluate the accuracy of the best models for each target related to the conventional methods, the RPD for the predictions were calculated (Table \ref{tab_rpd}). The standard deviation data used for RPD calculation is on the Appendix A. 

\begin{table}[h]
    \centering
\begin{tabular}{cccc}
\hline
\textbf{Targets} & \textbf{Best model(s)} & \textbf{RMSE} & \textbf{RPD }\\
\hline
P (mg kg$^{-1}$) & MTSG (RF) & 7.6 & 1.2 \\
TOC (g kg$^{-1}$) & MTSG (SVM\_R, RF) & 2.1 & 2 \\
pH & All models & 0.3 & 1.3 \\
H+Al (cmol$_c$ kg$^{-1}$) & Several models & 0.9 & 1.3 \\
Ca (cmol$_c$ kg$^{-1}$) & Several models & 0.8 & 1.9 \\
Mg (cmol$_c$ kg$^{-1}$) & All models & 0.4 & 1.5 \\
K (cmol$_c$ kg$^{-1}$) & MTSG (SVM\_R) & 0.09 & 1.3 \\
SB (cmol$_c$ kg$^{-1}$) & Several models & 1 & 1.9 \\
CEC (cmol$_c$ kg$^{-1}$) & MTAS (SVM\_L), MTSG (SVM\_R, RF) & 0.8 & 1.9 \\
BSP (\%) & DRS (RF) & 6.1 & 1.8 \\
\hline
\end{tabular} 
    \caption{RPD performance of the best models.}
    \label{tab_rpd}
\end{table}{}

The targets that presented poor prediction, where only high and low values are distinguishable, were P, pH, H+Al and K. These targets had low prediction performance with EDXRF sensor due to its low detection sensibility for light elements. For Mg, the models obtained fair predictions, indicating the viability of the method. BSP is in the threshold of fair and good predictions. Ca, SB and CEC presented good predictions. Finally, TOC is in the threshold between good and very good quantitative predictions.

\section{Conclusion}\label{sec_conclusion}

In this work, we evaluated the usage of MTR methods to predict 10 soil parameters based on EDXRF as input information. We also developed MTSG, a novel MTR method. In relation to the aRRMSE, MTR methods were able to reduce the error over ST, especially MTAS, DRS and MTSG. Concerning RPT, for all targets, all MTR methods were better than or equal to their respective regressor version in ST. 
MTSG  was the best method for predicting the soil characteristics of the problem, obtaining the lowest RMSE for 4 out of the 10 targets and was able to reduce the baseline aRRMSE from 0.67 to 0.64, representing a global improvement of 4.48\%. In the best target improvement, i.e. base saturation percentage (BSP), the proposed method boosted in 19\% the predictive performance using SVM\_R baseline. The comparison to reference methods also showed that the predictions were at least fair for 6 targets. Based on the results, MTR is capable of improving the prediction of soil properties. 

As future work, we will investigate the use of MTSG with more different base-learners when creating the base-models, even the meta-model, since it was observed suitable algorithms for particular targets. Besides, the combination of different types of PSS (e.g. Infrared spectroscopy, Raman spectroscopy, EDXRF, etc) and MTR methods may offer a viable solution for direct soil analysis, and it can be further explored.

\section{Acknowledgement }

This study was financed in part by the Coordenação de Aperfeiçoamento de Pessoal de Nível Superior - Brasil (CAPES) - Finance Code 001, CNPQ (grants \#420562/2018-4, \#142985/2016-3 and \#304722/2017-0), São Paulo Research Foundation (FAPESP) grant \#2018/07319-6 and Funda\c c\~ao Arauc\'aria (Brazilian Agencies).
We would like to thank  Dr. Graziela M.C. Barboza, Dr. José Francirlei de Oliveira and IBITIBA research project (32.02.120.00.00) for the support in sample collection and conventional analysis.

\newpage
\bibliographystyle{elsarticle-harv}
\bibliography{ref}

\newpage
\section*{Appendices}
\section*{Appendix A: General comparison of MTSG with other methods}

The Table \ref{tab_reference} presents the descriptive statistic of the soil parameters determined by conventional analysis for all samples, calibration set and prediction set.

\begin{table}[h]
    \centering
    \resizebox{0.85\textwidth}{!}{
\begin{tabular}{ccccc}
\hline
\textbf{Parameters} & \textbf{Mean $\pm$ SD} & \textbf{CV (\%)} & \textbf{Minimum} & \textbf{Maximum} \\
\hline
\textbf{All samples (n=396)} & & & & \\
TOC (g kg$^{-1}$) & 19.8 $\pm$ 4.1 & 22 & 7.6 & 30.5 \\
pH & 5.3 $\pm$ 0.4 & 8 & 4.2 & 6.5 \\
H+Al (cmol$_c$ kg$^{-1}$) & 5.4 $\pm$ 1.1 & 20 & 2.7 & 9.7 \\
Ca (cmol$_c$ kg$^{-1}$) & 6.0 $\pm$ 1.5 & 25 & 1.4 & 9.2 \\
Mg (cmol$_c$ kg$^{-1}$) & 2.0 $\pm$ 0.6 & 28 & 0.7 & 3.7 \\
K (cmol$_c$ kg$^{-1}$) & 0.24 $\pm$ 0.16 & 67 & 0.05 & 1.15 \\
P (mg kg$^{-1}$) & 13.9 $\pm$ 9.9 & 71 & 0.4 & 69.8 \\
SB (cmol$_c$ kg$^{-1}$) & 8.2 $\pm$ 1.9 & 23 & 2.1 & 12.6 \\
CEC (cmol$_c$ kg$^{-1}$) & 13.7 $\pm$ 1.5 & 11 & 9.3 & 17.9 \\
BSP (\%) & 60 $\pm$ 10 & 18 & 18 & 80 \\
\hline
\textbf{Calibration set (n=264)} & & & & \\
TOC (g kg$^{-1}$) & 20.0 $\pm$ 4.4 & 22 & 7.6 & 30.5 \\
pH & 5.3 $\pm$ 0.4 & 8 & 4.2 & 6.5 \\
H+Al (cmol$_c$ kg$^{-1}$) & 5.4 $\pm$ 1.0 & 19 & 2.7 & 9 \\
Ca (cmol$_c$ kg$^{-1}$) & 6.1 $\pm$ 1.5 & 24 & 1.4 & 9.2 \\
Mg (cmol$_c$ kg$^{-1}$) & 2.1 $\pm$ 0.6 & 28 & 0.7 & 3.5 \\
K (cmol$_c$ kg$^{-1}$) & 0.25 $\pm$ 0.17 & 68 & 0.05 & 1.15 \\
P (mg kg$^{-1}$) & 14.1 $\pm$ 10.2 & 72 & 0.4 & 69.8 \\
SB (cmol$_c$ kg$^{-1}$) & 8.4 $\pm$ 1.8 & 22 & 2.2 & 12.6 \\
CEC (cmol$_c$ kg$^{-1}$) & 13.8 $\pm$ 1.6 & 11 & 9.3 & 17.9 \\
BSP (\%) & 60 $\pm$ 9 & 15 & 20 & 80 \\
\hline
\textbf{Prediction set (n=132)} & & & & \\
TOC (g kg$^{-1}$) & 19.4 $\pm$ 4.2 & 22 & 9.5 & 29.2 \\
pH & 5.3 $\pm$ 0.4 & 8 & 4.2 & 6.5 \\
H+Al (cmol$_c$ kg$^{-1}$) & 5.5 $\pm$ 1.2 & 22 & 3.1 & 9.7 \\
Ca (cmol$_c$ kg$^{-1}$) & 5.6 $\pm$ 1.5 & 27 & 1.4 & 8.7 \\
Mg (cmol$_c$ kg$^{-1}$) & 2.0 $\pm$ 0.6 & 30 & 0.6 & 3.7 \\
K (cmol$_c$ kg$^{-1}$) & 0.20 $\pm$ 0.12 & 60 & 0.05 & 0.65 \\
P (mg kg$^{-1}$) & 13.5 $\pm$ 9.4 & 70 & 0.6 & 55 \\
SB (cmol$_c$ kg$^{-1}$) & 7.9 $\pm$ 1.9 & 24 & 2.1 & 11.9 \\
CEC (cmol$_c$ kg$^{-1}$) & 13.4 $\pm$ 1.4 & 11 & 9.4 & 17.2 \\
BSP (\%) & 58 $\pm$ 11 & 18 & 18 & 77 \\
\hline
\end{tabular} 
}
    \caption{Result of descriptive statistics of soil parameters determined by conventional methods. SD represents standard deviation and CV, coefficient of variation.
}

    \label{tab_reference}
\end{table}{}

\newpage

\section*{Appendix B: General comparison of MTSG with other methods}

MTSG is being first proposed in this work. To validate this method, MTSG performance was assessed in benchmarking datasets commonly used in MTR literature. 

These datasets enclose different kind of problems, as shown in \citep{Spyromitros2016}: air ticket prices (atp1d and atp7d),  machining parameter settings  (edm), solar flares types (sf1 and sf2), heavy metals concentration in soil (jura), energy efficient buildings requirements (enb), concrete properties (slump), water quality properties (andro) and online engagement (scpf).

The aRRMSE resulting of the usage of ST, SST, ERC, MTAS, MTSG, MOTC and DRS with the base learners RF and SVM\_R for these datasets are presented in Table \ref{tab_appendix}. The average rankings obtained by each of the MTR methods are also presented at the bottom of the table for visual clarity.

\begin{table}[h]
\centering
\begin{adjustbox}{width=\columnwidth,center}
\begin{tabular}{c|cc|cc|cc|cc|cc|cc|cc}
\hline
\multirow{2}{*}{\textbf{Dataset}} & \multicolumn{2}{c|}{ \textbf{ST} } & \multicolumn{2}{c|}{ \textbf{SST} } & \multicolumn{2}{c|}{ \textbf{ERC} } & \multicolumn{2}{c|}{ \textbf{MTAS} }  & \multicolumn{2}{c|}{ \textbf{MOTC }} & \multicolumn{2}{c|}{ \textbf{DRS} } & \multicolumn{2}{c}{ \textbf{MTSG} } \\
\cline{2-15}
 & RF & SVM\_R & RF & SVM\_R & RF & SVM\_R & RF & SVM\_R & RF & SVM\_R & RF & SVM\_R & RF & SVM\_R \\
\hline
atp1d & 0.3927 & 0.4291 & 0.3889 & 0.4290 & 0.3898 & 0.4290 & \textbf{0.3841} & \textbf{0.4170} & 0.4064 & 0.4627 & 0.3894 & 0.4311 & 0.3935 & 0.4230 \\
atp7d & 0.5108 & 0.6169 & 0.5041 & 0.6166 & 0.5094 & 0.6169 & 0.5107 & 0.5983 & 0.5844 & 0.7805 & \textbf{0.5025 } & 0.6159 & 0.5250 & \textbf{0.5877} \\
edm & 0.6668 & 0.7400 & 0.7146 & 0.7321 & \textbf{0.6620} & 0.7355 & 0.6852 & 0.6779 & 0.6655 & 0.8761 & 0.6799 & 0.7489 & 0.7404 & \textbf{0.6691} \\
sf1 & 0.8749 & 0.8054 & 1.0145 & 0.8131 & 0.9078 & \textbf{0.8013} & 1.0347 & 0.8167 & \textbf{0.8723} & 0.9938 & 0.8744 & 0.8040 & 1.0499 & 1.0001 \\
sf2 & \textbf{0.8254} & 0.7840 & 0.8955 & 0.7878 & 0.8399 & 0.7851 & 0.9150 & 0.7894 & 0.8280 & 1.5992 & 0.8303 & 0.\textbf{7829} & 0.9324 & 1.4815 \\
jura & 0.5842 & 0.6210 & 0.5734 & 0.6240 & 0.5752 & 0.6202 & \textbf{0.5673} & \textbf{0.5997} & 0.5792 & 0.6967 & 0.5758 & 0.6280 & 0.5739 & 0.6596 \\
enb & 0.1504 & 0.2510 & 0.1145 & 0.2190 & 0.1293 & 0.2415 & \textbf{0.1085} & \textbf{0.1236} & 0.1169 & 0.3171 & 0.1129 & 0.1688 & 0.1162 & 0.2017 \\
slump & 0.8293 & 0.7348 & 0.8484 & 0.7510 & 0.8182 & 0.7293 & 0.8159 & 0.8165 & \textbf{0.7633} & \textbf{0.7039} & 0.8654 & 0.7364 & 0.7982 & 0.7487 \\
andro & 0.8131 & 1.1912 & 0.7330 & 1.0197 & 0.7886 & 1.0929 & 0.5952 & 0.8980 & 0.5568 & 1.1806 & \textbf{0.5435} & 0.7713 & 0.6078 & \textbf{0.6009} \\
scpf & 0.8683 & 0.8190 & 0.8533 & \textbf{0.8016} & \textbf{0.8321} & 0.8060 & 0.9466 & 2.5352 & 0.8701 & 3.3978 & 0.8444 & 0.8026 & 0.8793 & 0.8033 \\

\hline
Average rank & 7.20 & 8.50 & 6.60 & 7.80 & 6.20 & 7.70 & 6.20 & 7.90  & 5.90 & 12.20 & \textbf{5.40} & 7.30 & 8.10 & 8.00\\

\hline
\end{tabular} 
    \end{adjustbox}
    \caption{aRRMSE comparison among the methods for literature benchmarking datasets. The bold values correspond to the smallest aRRMSE per dataset for each learning algorithm.}
    \label{tab_appendix}
\end{table}{}

As observed, none of the methods is the best for all the problems. MTAS presented the lowest performance in 3 datasets for both RF and SVM\_R. ERC, MTSG, MOTC and DRS presented the lowest aRRMSE for 3 datasets each, depending on the base regressor. SST and ST were able to minimise the aRRMSE once each. 

To conclude, the result shows that MTSG performance is comparable to previous proposed methods. Also, depending on the problem, its performance can be superior to the other methods.

\end{document}